\documentclass[11pt]{article}
\usepackage[utf8]{inputenc}
\usepackage{graphicx}
\usepackage{subfig}
\usepackage{amsfonts}
\usepackage{amsmath}
\usepackage{booktabs}
\usepackage{siunitx}
\usepackage{sansmath}
\usepackage{makecell,adjustbox}
\usepackage{url,hyperref,lineno,microtype}
\usepackage{epsfig,tikz,algpseudocode,algorithmicx,algorithm,pgfplots,geometry}
\usepackage{authblk}
\usetikzlibrary{bayesnet,fit,positioning}
\usepackage{siunitx}
\usepackage{sansmath} % for tensor notation
\DeclareMathAlphabet{\mathsfbfun}{\sansmathencoding}{\sfdefault}{bx}{n}
\newcommand{\tensor}[1]{\mathsfbfun{#1}}
\renewcommand*\vec[1]{\boldsymbol{\mathbf{#1}}}

\DeclareMathOperator{\Tr}{Tr}
\DeclareMathOperator{\E}{{\mathbb E}}

\DeclareMathOperator{\argmax}{arg\,max}

\DeclareMathOperator{\lse}{lse}

\DeclareMathOperator{\Cat}{Cat}
\DeclareMathOperator{\KL}{KL}
\DeclareMathOperator{\matvec}{vec}
\usepackage{natbib}
\begin{document}

\title{Factorisation-based Image Labelling}

\author[1,2]{Yu Yan}
\author[1,3]{Ya\"el Balbastre}
\author[1,2]{Mikael Brudfors}
\author[1,*]{John Ashburner}

\affil[1]{Wellcome Centre for Human Neuroimaging, UCL Queen Square Institute of Neurology, University College London, London, UK}
\affil[2]{School of Biomedical Engineering \& Imaging Sciences, King's College London, London, UK}
\affil[3]{Athinoula A. Martinos Center for Biomedical Imaging, Massachusetts General Hospital and Harvard Medical School, Boston, USA}
\affil[*]{Corresponding author.}
\date{}

\setcounter{Maxaffil}{0}
\renewcommand\Affilfont{\itshape\small}

\maketitle
\thispagestyle{empty}

\begin{abstract}
Segmentation of brain magnetic resonance images (MRI) into anatomical regions is a useful task in neuroimaging.
Manual annotation is time consuming and expensive, so having a fully automated and general purpose brain segmentation algorithm is highly desirable.
To this end, we propose a patched-based label propagation approach based on a generative model with latent variables.
Once trained, our Factorisation-based Image Labelling (FIL) model is able to label target images with a variety of image contrasts.
We compare the effectiveness of our proposed model against the state-of-the-art using data from the \emph{MICCAI 2012 Grand Challenge and Workshop on Multi-Atlas Labeling}.
As our approach is intended to be general purpose, we also assess how well it can handle domain shift by labelling images of the same subjects acquired with different MR contrasts.
\end{abstract}


%%%%%%%%%%%%%%%%%%%%%%%%%%%%%%%%%%%%%%%%%%%%%%%%%%%%%%%%%%%%%%%%%%%%%%%%%%%%%
\section{Introduction}

Accurate automated labelling of brain structures in MRI scans has many applications in neuroscience.
For example, studies of resting state fMRI or diffusion weighted imaging often involve summarising measures of connectivity among a relatively small number of brain regions.
Typically, these regions are only very approximately defined, using simple methods such aligning with a single manually labelled brain \citep{tzourio2002automated}.
While more accurate methods of brain parcellation are available, they usually have limitations in terms of the types of MRI scans that can be labelled, and they are often very computationally expensive.
For these reasons, they have not yet been widely adopted by the neuroimaging field.
We attempt to overcome some of these limitations by presenting a novel brain image labelling toolbox for the widely used SPM software.

In this work, we adopt a multi-atlas labelling approach.
Given a training set of scans from $N$ individuals, ${\bf X} = \{{\bf x}_1, {\bf x}_2, ... , {\bf x}_N\}$, along with their corresponding manual annotations ${\bf Y} = \{{\bf y}_1, {\bf y}_2, ... , {\bf y}_N\}$, the aim would be to estimate a suitable labelling $\hat{\bf y}^*$ for a target image ${\bf x}^*$ that is not part of the training set.
This objective is often achieved by obtaining the single most probable labelling, given by
\begin{align}
    \hat{\bf y}^* = \argmax_{{\bf y}^*} P({\bf y}^* | {\bf x}^*, {\bf X}, {\bf Y}).
\end{align}

Most label propagation methods require alignment between all the atlases in the training data (${\bf X}$) and the target image (${\bf x}^*$), which is usually achieved by a series of pairwise registrations.
The Symmetric Normalization algorithm (in the ANTS package) \citep{avants2008symmetric} is popular for this, although other algorithms are available.
This pair-wise strategy allows the label propagation to be performed directly in the space of the target image, which may lead to increased robustness because a small proportion of failures should have a relatively small impact on the results.

Once mappings between the training images and target image have been established, the manually defined labels are warped using the same mappings.
Each of these then provides a candidate labelling of the target image.
At this stage, some form of machine learning approach is used to predict the labels ($\hat{\bf y}^*$) for the target image from all the candidate labellings.
This procedure is often conceptualised in terms of some form of local weighted voting strategy, the simplest of which is to give each an equal vote.
In practice, this system is less effective than one that is weighted according to how well informed the voters are.

Earlier methods were similar to k-nearest neighbour classification, whereby a subset of atlases were chosen to label each brain region \citep{rohlfing2004evaluation}.
Others use a non-local patch-based framework \citep{coupe2010nonlocal}, with greater weighting for votes based on a more accurate model of the target image.
This typically involves a measure of similarity between patches in  ${\bf x}^*$ and the corresponding patches within each aligned image in ${\bf X}$.
Many such approaches can be conceptualised as a joint non-parametric generative model of both image and label data \citep{sabuncu2010generative}. 
An alternative framework is Simultaneous Truth and Performance Level Estimation (STAPLE) \citep{warfield2004simultaneous}, which is based on weighting the votes from candidate atlases according to how well they match the consensus over all votes.
More local versions of STAPLE have also been devised \citep{commowick2012estimating,asman2012formulating}, as well as hybrids between STAPLE and the non-local approaches \citep{asman2012multi}. 
Other approaches involve assuming that all voxels within each labelled region should have similar intensities, in a similar way to how many domain-adaptive tissue segmentations work.
Some methods treat this as a post-processing step in conjunction with a Markov random field (MRF) \citep{ledig2013improving}, whereas others have used this assumption to drive the image registration \citep{tang2013bayesian}.

Rather than conceptualise label propagation entirely in terms of optimising vote weightings, more recent work considers it within the more general pattern recognition framework.
This includes using Random Forest approaches \citep{zikic2014encoding} and, more recently, using convolutional neural networks \citep{de2015deep,moeskops2016automatic,mehta2017brainsegnet,roy2017error,wachinger2018deepnat,kushibar2019supervised,roy2019quicknat,rashed2020end}.

Many of these methods are impacted by the problem of domain shift, which is the situation where images in the training data (${\bf X}$) have different properties from those that the algorithm is to be applied to (${\bf x}^*$).
Typically, this is due to differences between image acquisition settings, scanner vendors, field strength, and so on.
Our aim is to release an automated labelling procedure for general purpose use, which would require overcoming the domain shift problem.
We attempt to circumvent it by working with images that have previously been segmented into different tissue types.
This can be achieved using one of the many domain-adaptive brain image segmentation approaches that have been developed.
Such approaches generally build on the idea of fitting some form of clustering model to the data \citep{wells1996adaptive,ashburner1997multimodal}, often with MRFs \citep{van1999automated,zhang2001segmentation} or deformable tissue priors \citep{ashburner2005unified,pohl2006bayesian,puonti2016fast} built in.
While reducing medical images to a few tissue types inevitably leads to some useful information being lost, we consider that this is a price worth paying for the increase in generality of the approach.
A related strategy (this time separating segmentation from diagnosis) has been used for increasing the generalisability of deep learning approach for diagnosing retinal disease \citep{de2018clinically}.

In our proposed Factorisation-based Image Labelling (FIL) method, we consider multi-atlas labelling as a special case of the more general problem of image-to-image translation, but where some of the data are binary or categorical in nature. Hence, our approach differs from previous methods in a number of ways.

Because running many pairwise registrations can be quite time consuming, we propose to label target images using only a single image registration.
Training involves first running an image registration approach to warp all training data (${\bf X}$) into the same atlas space (i.e., spatially normalise the data).
After this, only a single image registration step is required to align any target image with the average atlas space.
While we acknowledge that this may sacrifice some registration accuracy and robustness, we argue that working in a symmetric space (where no image is a \emph{source} or \emph{target}) facilitates pattern recognition across the set of training scans.

Other methods use purely discriminative methods for the label propagation itself.
These methods model ${\bf y}^*$ conditional on ${\bf x}^*$, whereas we propose to use a parametric generative model that encodes the joint probability of ${\bf x}^*$ and ${\bf y}^*$.
In addition to providing a new way of thinking about label propagation, we hope this generative model will open up other avenues of exploration in the future, particularly regarding multi-task and semi-supervised learning \citep{zhu2005semi}. 
%%%%%%%%%%%%%%%%%%%%%%%%%%%%%%%%%%%%%%%%%%%%%%%%%%%%%%%%%%%%%%%%%%%%%%%%%%%%%

%%%%%%%%%%%%%%%%%%%%%%%%%%%%%%%%%%%%%%%%%%%%%%%%%%%%%%%%%%%%%%%%%%%%%%%%%%%%%
\section{Methods}

The basic idea is to learn a generative latent variable model that can be used to project labels onto images that have been already warped into approximate alignment with each other (i.e., spatially normalised).
Alignment across individuals provides prior knowledge about the approximate locations of the various brain structures.
Therefore, the fully convolutional machinery of convolutional neural networks (CNNs) is not employed because it may not fully use this prior knowledge.

Rather than use the original pixel/voxel values, the approach aims to achieve generalisability across different types of images by working with categorical maps obtained from one of many tissue classification algorithms.
Some confusion may arise from the fact that both our \emph{images} and \emph{labels} are categorical. We will use \emph{categorical images} to denote the original images segmented into tissue types and \emph{categorical labels} to denote manually delineated anatomical labels. The number of anatomical classes is typically larger (and more fine-grained) than the number of tissue classes; although this is not a requirement of the model, which is symmetric with respect to both entries.

\begin{figure}[h!]
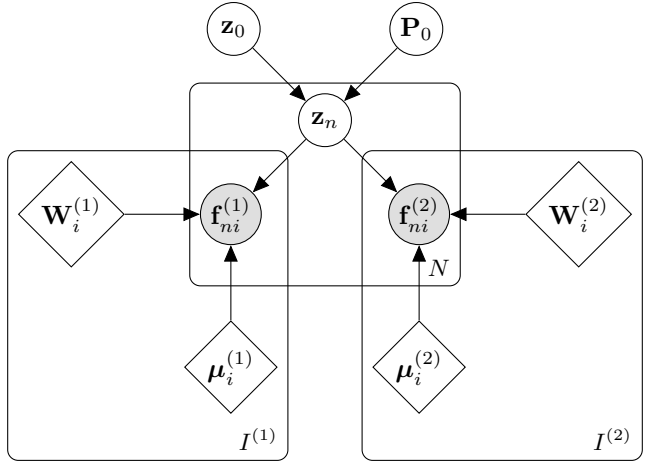

\begin{center}
\tikz{
    \node[latent                ] (z)   {${\bf z}_n$} ;
    \node[obs, below  left=of z ] (f1)  {${\bf f}^{(1)}_{ni}$} ;
    \node[obs, below right=of z ] (f2)  {${\bf f}^{(2)}_{ni}$} ;
    \node[det,       below=of f1] (mu1) {$\boldsymbol{\mu}^{(1)}_i$} ;
    \node[det,       below=of f2] (mu2) {$\boldsymbol{\mu}^{(2)}_i$} ;
    \node[det,        left=of f1] (W1)  {${\bf W}^{(1)}_i$} ;
    \node[det,       right=of f2] (W2)  {${\bf W}^{(2)}_i$} ;
    \node[latent, above right=of  z] (P0) {${\bf P}_0$} ;
    \node[latent, above  left=of  z] (z0) {${\bf z}_0$} ;
    \edge{mu1,W1,z}{f1};
    \edge{mu2,W2,z}{f2};
    \edge{z0,P0}{z};
    \plate{I1}{(f1)(W1)(mu1)}{$I^{(1)}$};
    \plate{I2}{(f2)(W2)(mu2)}{$I^{(2)}$};
    \plate{N}{(z)(f1)(f2)}{$N$};
}
\end{center}
\caption{Generative model for a single patch.
Data from $N$ subjects are assumed to be patches containing categorical data.
Image voxels are denoted by ${\bf f}^{(1)}_{ni}$ (where $n\in[1..N]$ and $i\in[1..I^{(1)}]$) and label voxels by ${\bf f}^{(2)}_{ni}$ ($i\in[1..I^{(2)}]$).
These are encoded by their means ($\boldsymbol{\mu}^{(1)}_i$ and $\boldsymbol{\mu}^{(2)}_i$ respectively) and a linear combination of basis functions (${\bf W}^{(1)}_i$ and ${\bf W}^{(2)}_i$ respectively).
For each subject, the contributions of the two sets of basis functions are jointly controlled by latent variables ${\bf z}_n$, which are assumed to be drawn from a normal distribution of mean ${\bf z}_0$ and precision ${\bf P}_0$.
\label{fig:PatchModel}}
\end{figure}

The approach is patch-based and applied to spatially normalised tissue maps (i.e., categorical image data). It can be seen as a multinomial version of principal component analysis, similarly to how linear regression can be generalised to multinomial logistic regression.
For each patch, a set of basis functions model both the categorical image data and the corresponding categorical label data, with a common set of latent variables controlling the contributions from the two sets of basis functions.
The results are passed through a softmax to encode the means of a multinomial distribution.
Training the model at a patch involves making point estimates of the set of spatial basis functions ($\tensor{W}^{(1)}$ and $\vec{M}^{(1)}$) that model the categorical image data ($\tensor{F}^{(1)}$), along with the basis functions ($\tensor{W}^{(2)}$ and $\vec{M}^{(2)}$) that model the label data ($\tensor{F}^{(2)}$).
For subject $n$, the contributions of the basis functions to both the image patch and its corresponding label patch are controlled by the same set of latent variables ($\vec{z}_n$).
The overall model for a patch is presented in Fig. \ref{fig:PatchModel}.

Once trained, the strategy is to determine the distribution of $\vec{z}^*$ for each patch within the target image data, which we refer to as the ``encoding step''.
This is achieved by fitting the learned $\tensor{W}^{(1)}$ and $\vec{M}^{(1)}$ to each patch ($\tensor{F}^{*(1)}$).
Then by using $\tensor{W}^{(2)}$ and $\vec{M}^{(2)}$, and given the estimated distribution of $\vec{z}^*$, it is possible to use a ``decoding step'' to probabilistically predict the unknown labels ($\tensor{F}^{*(2)}$) for the patch. 

The next section describes a simplified version of the full model shown in Fig. \ref{fig:PatchModel}.
Rather than model both $\tensor{F}^{(1)}$ and $\tensor{F}^{(2)}$, it describes how to encode a single patch $\tensor{F}$ using an approach similar to a generalisation of principal component analysis (PCA).
Generalising this simple approach to jointly model $\tensor{F}^{(1)}$ and $\tensor{F}^{(2)}$ is relatively straightforward.

\subsection{Multinomial Logistic Principal Component Analysis Model \label{Sec:MLPCA}}

This section describes a multinomial principal component analysis (PCA), which is based on \citet{khan2010variational}.
For PCA of categorical data, the arrays involved are multi-dimensional, so we represent them as collections of matrices at each of the $I$ voxels 

\begin{alignat}{3}
&\text{Categorical images: }~ &&\tensor{F} = && \left\{{\bf F}_i \in [0,1]^{M \times N} ~\middle|~ \sum_{m=1}^M f_{m n i} \in [0, 1] \right\}_{i=1}^I\\
&\text{Basis functions: }~ &&\tensor{W} = && \left\{{\bf W}_i \in \mathbb{R}^{M \times K}\right\}_{i=1}^I\\
&\text{Means: } &&{\bf M}~ = && \left\{\boldsymbol{\mu}_i \in \mathbb{R}^{M \times 1}\right\}_{i=1}^I.
\end{alignat}

Note that categories lie on the simplex: there are $M+1$ mutually exclusive categories within the data (i.e., the number of possible labels), but there is no need to represent all categories because the final category is determined by the initial $M$ categories.
Dimension $N$ denotes the number of items (i.e., the number of images in the training set).
Dimension $K$ denotes the number of basis functions in the model, and therefore also the number of latent variables per item
\begin{align}
{\bf Z} \in \mathbb{R}^{K \times N}.
\end{align}

Our notation uses a bold sans serif font to denote 3D tensors (e.g. $\tensor{F}$).
Each slice is a 2D matrix, shown in bold serif upper case (e.g. ${\bf F}_i$).
The next level of indexing extracts column vectors from matrices, which are shown in bold lower case (e.g. ${\bf f}_{ni}$).
Dimensions are shown in upper case italic (e.g. $N$), whereas indices and other scalars are in lower case italic (e.g. $n$ or $f_{mni}$).
This work considers training a generalised PCA model to find the most probable values of ${\bf M}$ and $\tensor{W}$, while marginalising with respect to ${\bf Z}$
\begin{align}
\hat{\bf M},\hat{\tensor{W}} ={} & \argmax_{{\bf M},\tensor{W}} p(\tensor{F},\tensor{W} | {\bf M})\\
p(\tensor{F},\tensor{W}|{\bf M}) ={} & \int_{\bf Z} P(\tensor{F}|{\bf Z},\tensor{W},{\bf M}) p({\bf Z}) p(\tensor{W}) d{\bf Z}.
\end{align}
Within our model, the basis functions ($\tensor{W}$) are assumed to be drawn from
\begin{align}
    {\bf w}_{ki} \sim \mathcal{N}\left({\bf 0}, \left({\bf I}_M + \tfrac{1}{M+1}\right)^{-1}\right).
\end{align}

No priors are imposed on the mean (${\bf M}$).
The latent variables are assumed to be drawn from an empirically determined (see Section \ref{sec:CRF}) Gaussian distribution that imposes spatial contiguity between neighbouring patches, such that
\begin{align}
p({\bf Z}) ={} & \prod_{n=1}^N p({\bf z}_n)\\
    {\bf z}_n \sim{} & \mathcal{N}({\bf z}_0, {\bf P}_0^{-1}). \label{eqn:pz}
\end{align}

In linear PCA, the likelihood takes the form of an isotropic Gaussian distribution conditioned on the reconstructed data ($\vec{\eta}_{ni} = \vec{\mu}_{i} + \vec{W}_{i}\vec{z}_{n}$). Here, it is based around a categorical distribution conditioned on the \emph{softmaxed} reconstruction, similar to frameworks used for multinomial logistic regression
\begin{align}
P(\tensor{F} | {\bf Z}, \tensor{W}, {\bf M}) ={} & 
\prod_{i=1}^I \prod_{n=1}^N P({\bf f}_{ni} | {\bf z}_n, {\bf W}_i, \boldsymbol\mu_i) \\
P({\bf f}_{ni} | {\bf z}_n, {\bf W}_i, \boldsymbol\mu_i) ={} & P({\bf f}_{ni} | \boldsymbol\rho_{ni}) = \prod_{m=1}^{M} \rho_{mni}^{f_{mni}}. 
\label{eqn:Cat}
\end{align}

The mean (${\boldsymbol\rho}_{ni}$) of each categorical distribution is computed as the softmax ($\sigma$) of the linear combination of basis functions
\begin{align}
{\boldsymbol\rho}_{ni} = & \sigma\left({\boldsymbol\eta}_{ni}\right) = \frac{\exp({\boldsymbol\eta}_{ni})}{1+\sum_{m=1}^M\exp(\eta_{mni})} \label{eqn:softmax}\\
{\boldsymbol\eta}_{ni} = & {\bf W}_i {\bf z}_n + \boldsymbol{\mu}_i.
\end{align}

The log-likelihood from Eqs. \ref{eqn:Cat} and \ref{eqn:softmax} is re-written to make use of the log-sum-exp ($\lse$) function
\begin{align}
\ln P({\bf f}_{ni} | {\boldsymbol\eta}_{ni}) = & \boldsymbol\eta_{ni}^T {\bf f}_{ni} - \lse(\boldsymbol\eta_{ni}) \label{Eq:loglik}\\
\text{where}\cr
\lse(\boldsymbol\eta_{ni}) = & \ln \left(1 + \sum_{m=1}^M \exp(\eta_{mni}) \right).
\end{align}

Numerous computations within this work benefit from having a local quadratic approximation to the $\lse$ function around some point $\hat{\boldsymbol\eta}_{ni} = {\bf W}_i \hat{\bf z}_n + \boldsymbol{\mu}_i$.
This is based on a second order Taylor polynomial that uses the gradient ($\hat{\boldsymbol\rho} = \sigma\left(\hat{\boldsymbol\eta}_{ni}\right)$), but replaces the true Hessian with B\"ohning's approximation (${\bf A}$) \citep{bohning1992multinomial}
\begin{align}
\lse(\boldsymbol\eta_{ni})
\le & \lse(\hat{\boldsymbol\eta}_{ni}) + (\boldsymbol\eta_{ni}-\hat{\boldsymbol\eta}_{ni})^T \hat{\boldsymbol\rho}_{ni} + \tfrac{1}{2}(\boldsymbol\eta_{ni}-\hat{\boldsymbol\eta}_{ni})^T {\bf A} (\boldsymbol\eta_{ni}-\hat{\boldsymbol\eta}_{ni}) \label{eqn:Bohning_approx}\\
\text{where}\cr
\hat{\boldsymbol\rho}_{ni} = & \sigma(\hat{\boldsymbol\eta}_{ni}) = \exp(\hat{\boldsymbol\eta}_{ni} - \lse(\hat{\boldsymbol\eta}_{ni}))\\
{\bf A} = & \tfrac{1}{2}\left( {\bf I}_M - \tfrac{1}{M+1} \right).
\end{align}

\begin{figure}[h!]
\begin{center}
\epsfig{file=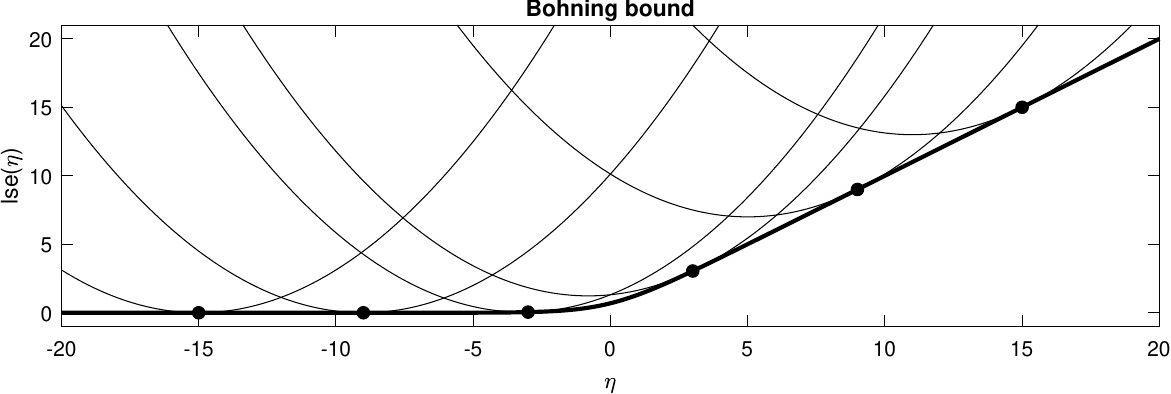,width=\textwidth}\\
\end{center}
\caption{
B\"ohning's bound plotted for various values of $\hat{\boldsymbol\eta}$ for the binary case.
The thick line shows the actual $\lse$ function, whereas the fine lines show the local quadratic approximations centred around various points (solid circles).
\label{Fig:bb}}
\end{figure}

It is worth noting that B\"ohning's proposed approximation is more positive definite (in the Loewner ordering sense) than any of the actual Hessians, which guarantees that the approximating function provides an upper bound to the true lse function \citep{bohning1992multinomial,khan2010variational} (see  Fig. \ref{Fig:bb} for an illustration where $M$=1).

\subsection{Model Training\label{Sec:training}}

This section describes how the basis functions are estimated for each patch.
A variational expectation maximisation (EM) approach is used for fitting the model in such a way that the uncertainty of ${\bf Z}$ can be accounted for.
Variational Bayesian methods are a family of technique for approximating the types of intractable integrals often encountered in Bayesian inference.
Further explanations may be found in textbooks, such as \citet{bishop2006pattern} and \citet{murphy2012machine}.
In summary, it uses an approximating distribution to enable a lower bound on a desired log-likelihood to be sought.
In this work, the approximating distribution is $q({\bf Z})$, which is used to provide a lower bound $\mathcal{L}(q)$ on the likelihood $P(\tensor{F}|{\bf M},\tensor{W})$.
This bound is tightest when $q({\bf Z})$ most closely approximates $p({\bf Z} | \tensor{F}, {\bf M}, \tensor{W})$ according to the Kullback-Liebler divergence (i.e. $\KL(q||p)$)
\begin{align}
\ln p(\tensor{F},\hat{\tensor{W}}| \hat{\bf M}) = & \mathcal{L}(q) + \KL(q||p)\\
\mathcal{L}(q) = &  \int_{\bf Z} q({\bf Z}) \ln \left( \frac{p(\tensor{F},{\bf Z},\hat{\tensor{W}} | \hat{\bf M})}{q({\bf Z})} \right) d{\bf Z} \label{eq:elbo}\\
\KL(q||p)      = & -\int_{\bf Z} q({\bf Z}) \ln \left( \frac{p({\bf Z} | \tensor{F}, \hat{\tensor{W}},\hat{\bf M})}{q({\bf Z})} \right) d{\bf Z}.
\end{align}

Fitting the model by variational EM involves an iterative algorithm that alternates between a variational E-step that updates the approximation to the distribution of the latent variables $q({\bf Z})$ using the current point estimates of ${\bf M}$ and ${\tensor{W}}$ (i.e., $\hat{\bf M}$ and $\hat{\tensor{W}}$), and a variational M-step that uses $q({\bf Z})$ to update $\hat{\bf M}$ and $\hat{\tensor{W}}$.
These two steps are outlined next.

\subsubsection{Variational E-step}

We choose a product of Gaussian distributions for the approximate latent distribution, such that each factor is parameterised by a mean $\hat{\vec{z}}_n$ and a covariance matrix $\vec{V}_n$
\begin{align}
    q(\vec{Z}) = \prod_{n=1}^N \mathcal{N}({\bf z}_n | \hat{\bf z}_n, {\bf V}_n) .
\end{align}

Keeping only terms that depend on $(\hat{\vec{z}}_n, \vec{V}_n)$, the evidence lower bound in Eq. \eqref{eq:elbo} can be written as
\begin{align}
    \mathcal{L}(\hat{\vec{z}}_n, \vec{V}_n) = \int_{\vec{z}_n} \left( \ln p({\bf F}_n, {\bf z}_n, \hat{\tensor{W}}| \hat{\bf M}, {\bf z}_0, {\bf P}_0)  -  \ln q(\vec{z}_n) \right) q(\vec{z}_n) d\vec{z}_n 
    + \text{const}.
\end{align}

Making use of the local approximation in Eq. \ref{eqn:Bohning_approx} about $\hat{\boldsymbol\eta}_{ni} = \hat{\bf W}_i \hat{\vec{z}}_n^{\text{prev}} + \hat{\boldsymbol\mu}_i$ and of Jensen's inequality, we can devise a quadratic lower bound on the evidence lower bound. Note that, as in classical EM, this quadratic lower-bound touches the variational lower-bound in each estimate; increasing the former therefore ensures to also increase the latter. A closed-form solution to the substitute problem exists, and we find
\begin{align}
{\bf V}_n     = & \left({\bf P}_0 + \sum_{i=1}^I \hat{\bf W}_{i}^T {\bf A} \hat{\bf W}_{i} \right)^{-1} \label{eqn:V}\\
\hat{\bf z}_n = & {\bf V}_n \left( {\bf P}_0 {\bf z}_0 + \sum_{i=1}^I \hat{\bf W}_i^T \left( {\bf f}_{ni} - \hat{\boldsymbol\rho}_{ni} + {\bf A}(\hat{\boldsymbol\eta}_{ni}-\hat{\boldsymbol\mu}_i)\right) \right) \label{eqn:hatZ}.
\end{align}

This Gaussian approximation has the following expectations, which can be substituted into various other equations when required
\begin{align}
\E[{\bf z}_n] = & \hat{\bf z}_n \label{eqn:Ez}\\
\E[{\bf z}_n {\bf z}_n^T] = & \hat{\bf z}_n \hat{\bf z}_n^T + {\bf V}_n. \label{eqn:Ezz}
\end{align}

\subsubsection{Variational M-step}

The M-step uses $q({\bf Z})$ to update the point estimates of ${\bf M}$ and $\tensor{W}$.
For simplicity, our implementation updates ${\bf M}$ and $\tensor{W}$ separately, although it would have been possible to update them simultaneously.
The strategy for updating $\hat{\bf M}$ is similar to a Gauss-Newton update, but we formulate it in a manner that would be familiar to those working with variational Bayesian methods.
This involves using the estimates of $\hat{\bf M}$, $\hat{\tensor{W}}$ and $\hat{\bf Z}$ to set the variational parameters to $\hat{\boldsymbol\eta}_{ni} = \hat{\bf W}_i \hat{\bf z}_n + \hat{\boldsymbol\mu}_i$.
Then the local approximation of Eq. \ref{eqn:Bohning_approx} is substituted for $\lse(\hat{\bf W}_i \hat{\bf z}_n + {\boldsymbol\mu}_i)$ in the expectation of $\ln p({\bf F}_i,{\bf Z}, \hat{\bf W}_i|\boldsymbol{\mu}_i)$ with respect to $q({\bf Z})$.
Terms that do not involve ${\boldsymbol{\mu}_i}$ are ignored, giving
\begin{align}
\E_{q({\bf Z})}[&\log p({\bf F}_i, {\bf Z}, \hat{\bf W}_i|\boldsymbol{\mu}_i)] 
\le \cr
&\boldsymbol\mu_i^T \sum_{n=1}^N({\bf f}_{ni} - \hat{\boldsymbol\rho}_{ni} + {\bf A}(\hat{\boldsymbol\eta}_{ni}-\hat{\bf W}_i \E[{\bf z}_n]) )
 - \tfrac{N}{2} \boldsymbol\mu_i^T {\bf A} \boldsymbol\mu_i + \text{const}.
\end{align}

Completing the square shows that $\boldsymbol\mu_i$ would be drawn from a multivariate Gaussian distribution, although we are only interested in its mean.
Substituting Eq. \ref{eqn:Ez} gives the following update for $\hat{\boldsymbol{\mu}}_i$
\begin{equation}
\hat{\boldsymbol{\mu}}_i = \left(N {\bf A}\right)^{-1} \left(\sum_{n=1}^N({\bf f}_{ni} - \hat{\boldsymbol\rho}_{ni} + {\bf A}(\hat{\boldsymbol\eta}_{ni}-\hat{\bf W}_i \hat{\bf z}_n))\right).
\end{equation}

A similar approach is used for updating ${\bf W}_i$, although the $\matvec$ operator is required to treat the ${\bf W}_i$ matrices as vectors of parameters.
The Kronecker tensor product ($\otimes$) is also used to construct the following upper bound, which is substituted into $\E_{q({\bf Z})}[\log p({\bf F}_i,{\bf Z}, {\bf W}_i|\hat{\boldsymbol{\mu}}_i)]$
\begin{align}
\E_{{\bf z}_n}[&{\bf f}_{ni}^T ({\bf W}_{i} {\bf z}_n + \hat{\boldsymbol\mu}_i) - \lse({\bf W}_{i} {\bf z}_n + \hat{\boldsymbol\mu}_i)]\cr
\le & -\tfrac{1}{2} \matvec({\bf W}_{i})^T (\E[{\bf z}_n {\bf z}_n^T] \otimes {\bf A}) \matvec({\bf W}_{i}) \cr
   & + \matvec({\bf W}_{i})^T ( \E[{\bf z}_n] \otimes ({\bf f}_{ni} - \hat{\boldsymbol\rho}_{ni} + {\bf A}(\hat{\boldsymbol\eta}_{ni} - \hat{\boldsymbol\mu}_i)))
+ \text{ const}.
\end{align}

Substituting Eqs. \ref{eqn:Ez} and \ref{eqn:Ezz} into $\E_{q({\bf Z})}[\log p({\bf F}_i,{\bf Z}, {\bf W}_i|\hat{\boldsymbol{\mu}}_i)]$ and completing the square reveals the following Gaussian distribution, and update for ${\bf W}_i$
\begin{align}
\matvec({\bf W}_{i}) \sim & \mathcal{N}(\matvec(\hat{\bf W}_{i}), {\bf H}_i^{-1})\\
    {\bf H}_i = & \sum_{n=1}^N (\hat{\bf z}_n \hat{\bf z}_n^T + {\bf V}_n) \otimes {\bf A} + {\bf I}_K \otimes ({\bf I}_M+\tfrac{1}{1+M})\\
\matvec(\hat{\bf W}_{i}) = & {\bf H}_i^{-1} \left(\sum_{n=1}^N \hat{\bf z}_n \otimes ({\bf f}_{ni} - \hat{\boldsymbol\rho}_{ni} + {\bf A}(\hat{\boldsymbol\eta}_{ni} - \hat{\boldsymbol\mu}_i)))\right).
\end{align}

\subsubsection{Conditional Random Field \label{sec:CRF}}

\begin{figure}
\begin{center}
\epsfig{file=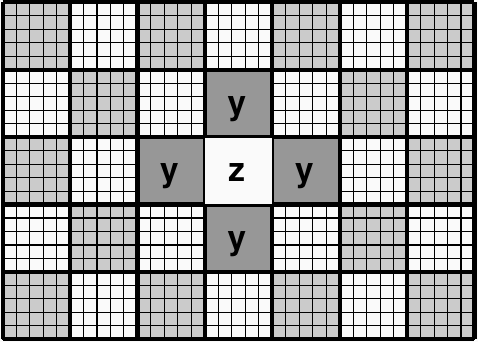,width=.5\textwidth}\\
\end{center}
\caption{
Schematic of the ``red-black'' checkerboard scheme in 2D.  In this illustration, updating the priors for the latent variables encoding the central (white) patch makes use of the latent variables from the four neighbouring (grey) patches. This illustrates that the latent variables in each patch are conditional on those of the neighbouring patches (Markov blanket).
\label{Fig:redblack}}
\end{figure}

This section describes how spatial contiguity is achieved.
Rather than treat the data as a collection of independent patches, the proposed approach attempts to model the relationship between the latent variables encoding each patch and those encoding the six immediately neighbouring patches (or four neighbouring patches in 2D).
For a valid mean-field approximation, updating patches is done via a ``red-black'' (checkerboard) ordering scheme (see Fig. \ref{Fig:redblack}), such that one pass over the data updates the ``red'' patches, while making use of the six neighbouring patches of each (which would correspond to ``black'' patches).
The next pass would update the ``black'' patches, while making use of the six (``red'') neighbouring patches of each.
The remainder of this section explains how information from neighbouring patches is used to provide empirical priors (${\bf z}_0$ and ${\bf P}_0$) for the latent variables of a central patch. 
We note that this approach is related to work by \citet{zheng2015conditional} and \citet{brudfors2019nonlinear}.

Recall that the posterior means and covariances of each latent variable are denoted $\hat{\bf z}_n$ and ${\bf V}_n$.
Here, we refer to the concatenated means and covariances of the latent variables in all adjacent patches as $\hat{\bf y}_n$ and ${\bf U}_n$ respectively, where ${\bf U}_n$ is block diagonal.
We assume that
\begin{align}
\left[\begin{matrix}
{\bf z}_n\cr
{\bf y}_n\cr
\end{matrix} \right] \sim \mathcal{N}({\bf 0},{\bf P}^{-1}).
\end{align}

Using $K$ to denote the order of ${\bf P}$, the model assumes a Wishart prior on ${\bf P}$.
\begin{align}
\ln p({\bf P})
= & \ln \mathcal{W}(\boldsymbol\Psi_0, \nu_0)\cr
= &  \tfrac{\nu_0-K-1}{2} \ln \det |{\bf P}| - \tfrac{1}{2} \Tr(\boldsymbol\Psi_0^{-1} {\bf P})\cr
  & -\tfrac{\nu_0}{2} \ln \det|\boldsymbol\Psi_0| - \tfrac{\nu_0 K}{2} \ln 2 - \ln \Gamma_K\left(\tfrac{\nu_0}{2}\right)
\end{align}

This leads to the following approximating distribution for ${\bf P}$.
\begin{align}
\ln q({\bf P}) = & \E_{q({\bf Z}),q({\bf Y})}[\ln p({\bf Z},{\bf Y} | {\bf P}) + \ln p({\bf P})]\cr
= & \tfrac{N}{2} \ln \det | {\bf P} | - \tfrac{1}{2}\Tr\left(\E\left[\sum_{n=1}^N [\begin{matrix}{\bf z}_n^T & {\bf y}_n^T\end{matrix}]^T [\begin{matrix}{\bf z}_n^T & {\bf y}_n^T\end{matrix}]\right] {\bf P} \right)\cr
  & + \tfrac{\nu_0-K-1}{2} \ln \det |{\bf P}| - \tfrac{1}{2} \Tr(\boldsymbol\Psi_0^{-1} {\bf P}) + \text{const}
\end{align}

By substituting expectations from Eqs. \ref{eqn:Ez} and \ref{eqn:Ezz}, we can represent this as the Wishart distribution
\begin{align}
    q({\bf P}) = & \mathcal{W}({\bf P} | \boldsymbol\Psi, \nu)
\end{align}
where
\begin{align}
\boldsymbol\Psi = & \left(\sum_{n=1}^N\left[
\begin{matrix}
\hat{\bf z}_n \hat{\bf z}_n^T + {\bf V}_n & \hat{\bf z}_n \hat{\bf y}_n^T                \cr
\hat{\bf y}_n \hat{\bf z}_n^T                 & \hat{\bf y}_n \hat{\bf y}_n^T + {\bf U}_n\cr
\end{matrix}
\right] + \boldsymbol\Psi_0^{-1} \right)^{-1}\\
\nu = & N + \nu_0.
\end{align}

From the properties of Wishart distributions, we have
\begin{align}
\E[{\bf P}] = \nu \boldsymbol\Psi. \label{eqn:EP}
\end{align}

For the next step, the matrices ${\bf P}$ and $\boldsymbol\Psi$ are conformably decomposed into
\begin{align}
{\bf P} = \left[\begin{matrix}
{\bf P}_{zz} & {\bf P}_{zy}\cr
{\bf P}_{yz} & {\bf P}_{yy}\cr
\end{matrix} \right] \text{ and }
\boldsymbol\Psi = \left[ \begin{matrix}
\boldsymbol\Psi_{zz} & \boldsymbol\Psi_{zy}\cr
\boldsymbol\Psi_{yz} & \boldsymbol\Psi_{yy}\cr
\end{matrix} \right].
\end{align}

Now, we wish to compute priors for the latent variables of each patch, conditional on the latent variables of the neighbours.
This can be derived from
\begin{align}
\E_{q({\bf y}_n),q({\bf P})}\left[\ln p({\bf z}_n,{\bf y}_n)\right]
= & -\tfrac{1}{2} \E_{q({\bf y}_n),q({\bf P})}\left[ [\begin{matrix}{\bf z}_n^T & {\bf y}_n^T\end{matrix}] {\bf P} [\begin{matrix}{\bf z}_n^T & {\bf y}_n^T\end{matrix}]^T\right] + \text{const}\cr
= & -\tfrac{1}{2} {\bf z}_n^T \E[{\bf P}_{zz}] {\bf z}_n - {\bf z}_n^T \E[{\bf P}_{zy}] \E[{\bf y}_n] + \text{const}.
\end{align} 

By completing the square and substituting expectations from Eqs. \ref{eqn:Ez} and \ref{eqn:EP}, we can derive suitable priors for use in Eq. \ref{eqn:pz} from Section \ref{Sec:MLPCA}.
\begin{align}
    {\bf z}_n \sim & \mathcal{N}({\bf z}_0, {\bf P}_0^{-1})\\
    \text{where}\cr
    {\bf P}_0 = & \E[{\bf P}_{zz}] = \nu \boldsymbol\Psi_{zz}. \label{eqn:P0}\\
    {\bf z}_0 = & -\E[{\bf P}_{zz}^{-1} {\bf P}_{zy} {\bf y}_n] = -\boldsymbol\Psi_{zz}^{-1} \boldsymbol\Psi_{zy} \hat{\bf y}_n \label{eqn:z0}.
\end{align}

\subsubsection{Implementation Details}

Training the model is an iterative procedure.
Our implementation consists of a number of outer iterations, each involving a ``red'' and ``black'' sweep through the data.
During each outer iteration, patches are updated by first determining a new ${\bf P}_0$ and a new prior expectation ${\bf z}_0$ for each ${\bf z}_n$.
Then from these priors, the variational EM steps are repeated five times within each patch.
For each of these sub-iterations, the E-step is run five times, as is the M-step.

Unlike most other methods, our proposed approach performs label propagation on spatially normalised versions of the images.
To account for this, our implementation considers the expansions and contractions involved in warping the images using the Jacobian determinants to weight the data appropriately, which is effected by a slight modification to the likelihood term in Eq. \ref{Eq:loglik}.
This weighting essentially is an integration by substitution, and is used both during the training and testing phases.

Because the method is patch-based, not every patch needs to encode all possible brain structure labels.
To save memory and computation, the model is set up so that each patch only encodes the categories that it requires.
This is determined by whether that category exists in the corresponding patch in all training scans, and results in dimension $M^{(2)}$ varying across patches.

The Bayesian formulation of the model tends towards an automatic relevance determination solution, whereby the distributions of some latent variables approach a delta function at zero.
To speed up the computations, the model is ``pruned'' after every second iteration of the training.
This involves using PCA to make $\hat{\bf Z}^T \hat{\bf Z}$ orthonormal within each patch (along with applying the corresponding rotations to each $\hat{\bf W}_i^{(1)}$, $\hat{\bf W}_i^{(2)}$ and ${\bf V}_n$).
Any latent variables that contributed a negligible amount to the model fit were then removed, along with the basis functions they controlled.

As a simple attempt to make better use of the limited number of labelled images, the training data are augmented by also using versions translated by integer numbers of voxels along all three directions, up to some maximum radius.
We weight the contribution of each presentation of the data according to the amount of translation used.
This weighting is based on Gaussian function of distance ($\exp(-\tfrac{1}{2} d^2/s^2)$, and is parameterised by a standard deviation ($s$).
The weights are re-normalised to sum to one over all possible amounts of translation.
Weighting enters into the algorithm as a modification to the updates in the variational M-step, as well as when updating the parameters of the conditional random field.

\subsection{Labelling a Target Image \label{Sec:applying}}

\begin{figure}
\begin{center}
\includegraphics[width=\textwidth]{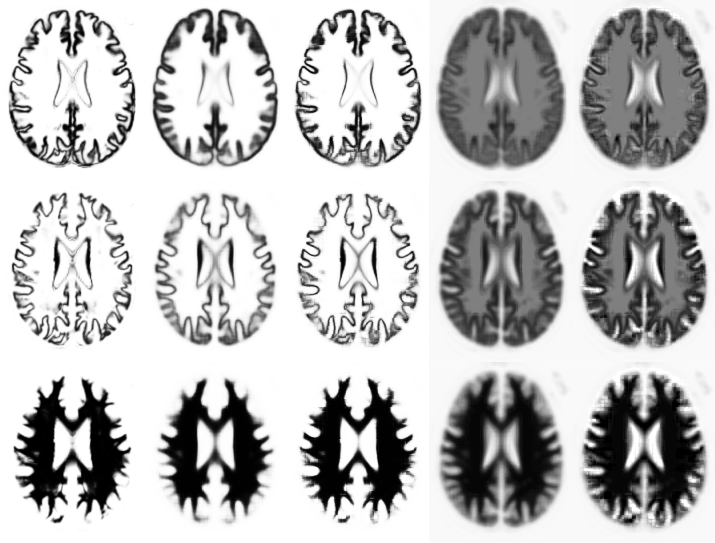}\\
\end{center}
\caption{
Illustration of (encoding) model fit to an example image, where each row shows a different tissue class.
The first column shows the warped categorical image data for one subject.
The second and fourth columns shows the mean (parameter ${\bf M}$ within each patch), with and without a softmax respectively.
The third and fifth columns show the model fit (mean plus linear combination of basis functions within each patch) to the categorical image data, with and without a softmax.  
\label{Fig:template_recon}}
\end{figure}

In this section, we explain the computations that take place during deployment of a trained model, such that the model can be used for labelling new and unseen images.
To re-iterate, labelling a new image involves an encoding step, where the distribution of $\vec{z}^*$ is estimated for each target image patch ($\mathbf{F}^{*(1)}$) by fitting $\vec{M}^{(1)}$ and $\tensor{W}^{(1)}$ to it (see Fig. \ref{Fig:template_recon}). This is followed by a decoding step, whereby the label probabilities ($\mathbf{F}^{*(2)}$) are reconstructed from the distribution of $\vec{z}^*$ using $\vec{M}^{(2)}$ and $\tensor{W}^{(2)}$.
We show that encoding a patch in a new image can be achieved by expressing Eqs. \ref{eqn:hatZ} and \ref{eqn:z0} as a type of recurrent ResNet \citep{he2016deep}.

\subsubsection{Encoding}

Latent variables are all initialised to zero, before the ``red-black'' scheme is used to update them.
One cycle updates the estimates of the latent variables ($\hat{\bf z}$) for the ``red'' patches, based on values of the latent variables in the neighbouring ``black'' patches ($\hat{\bf y}$).
The next cycle updates the latent variables of the ``black'' patches, using the recently updated neighbouring ``red'' latent variables.
Our implementation repeats this procedure for a fixed number of iterations, although it would be possible to terminate based on some convergence criterion.
For each patch, the parameters computed during training that are required during encoding are $\hat{\tensor{W}}^{(1)}$, ${\bf M}^{(1)}$, $\nu$ and $\boldsymbol\Psi$.
The procedure for computing the distribution of the latent variables ${\bf z}$ for a patch in a target image ${\bf F}^{*(1)} \in \{0,1\}^{M \times I}$ can be made more efficient by first pre-computing some new matrices.
The following covariance matrix is required, which is obtained by combining \ref{eqn:V} and \ref{eqn:P0}.
\begin{align}
{\bf V}     = & \left(\boldsymbol\Psi_{zz} \nu + \sum_{i=1}^I \hat{\bf W}_{i}^T {\bf A} \hat{\bf W}_{i} \right)^{-1}
\end{align}

The spatial basis functions ($\hat{\tensor{W}}^{(1)}$ and ${\bf M}^{(1)}$) are reshaped to make them easier to work with.
\begin{align}
{\bf W} = & \left[\begin{matrix}
\hat{\bf w}_{1 1}^{(1)} & \hat{\bf w}_{2 1}^{(1)} & \hdots & \hat{\bf w}_{K 1}^{(1)} \\
\hat{\bf w}_{1 2}^{(1)} & \hat{\bf w}_{2 2}^{(1)} & \hdots & \hat{\bf w}_{K 2}^{(1)} \\
\vdots        & \vdots        & \ddots & \vdots \\
\hat{\bf w}_{1 I}^{(1)} & \hat{\bf w}_{2 I}^{(1)} & \hdots & \hat{\bf w}_{K I}^{(1)}
\end{matrix}\right]
\text{ and }
\boldsymbol{\mu} = \left[\begin{matrix}
\boldsymbol{\mu}_1^{(1)}\\
\boldsymbol{\mu}_2^{(1)}\\
\vdots \\
\boldsymbol{\mu}_I^{(1)}
\end{matrix}\right]
\end{align}

The additional matrices that are pre-computed to speed up the updates in \ref{eqn:hatZ} are
\begin{align}
{\bf B}^{(0)} = & -{\bf V} \boldsymbol\Psi_{zy} \nu\\
{\bf B}^{(1)} = &  {\bf V} {\bf W}^T\\
{\bf B}^{(2)} = &  {\bf V} {\bf W}^T \left({\bf I}_I \otimes {\bf A} \right) {\bf W}.
\end{align}

Each patch in the target image is reshaped to a column vector.
\begin{align}
{\bf f}^* = & \left[\begin{matrix}
{\bf f}^{*(1)}_{1} \\
{\bf f}^{*(1)}_{2} \\
\vdots \\
{\bf f}^{*(1)}_{I}
\end{matrix}\right]
\end{align}

Computing the distribution of the latent variables can then be achieved by iterating the following.
\begin{align}
\hat{\bf z}^* \gets & {\bf B}^{(1)} {\bf f}^* + {\bf B}^{(0)} \hat{\bf y}^* + {\bf B}^{(2)} \hat{\bf z}^* - {\bf B}^{(1)} \sigma({\bf W} \hat{\bf z}^* + \boldsymbol{\mu} )
\end{align}

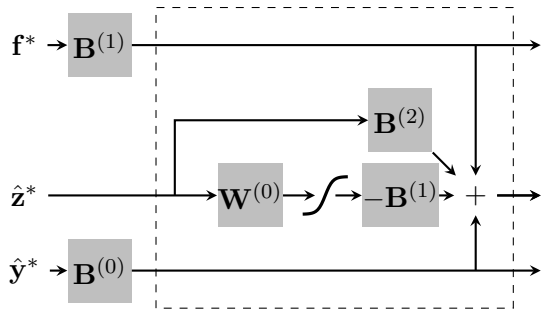
\begin{figure}[h!]
\begin{center}
    \begin{tikzpicture}[node distance=4cm]
    \tikzstyle{every pin edge}=[<-,shorten <=1pt]
    \tikzstyle{neuron}=[rectangle,fill=black!25,minimum size=24pt,inner sep=0pt]
    \tikzstyle{annot} = [text width=4em, text centered]
    \tikzstyle{arrow} = [thick,->,>=stealth]

    \node[]       (f)  at (-2, 0) {${\bf f}^*$};
    \node[neuron] (f1) at (-1, 0) {${\bf B}^{(1)}$};
    \node[]       (fo) at ( 5, 0) {};

    \node[]       (y)  at (-2,-3) {$\hat{\bf y}^*$};
    \node[neuron] (y1) at (-1,-3) {${\bf B}^{(0)}$};
    \node[]       (yo) at ( 5,-3) {};

    \node[]       (z)  at (-2,-2) {$\hat{\bf z}^*$};
    \node[neuron] (B2) at ( 3,-1) {${\bf B}^{(2)}$};
    \node[neuron] (z2) at ( 1,-2) {${\bf W}^{(0)}$};
    \node[]       (z3) at ( 2,-2) {};
    \draw[very thick] (0,0) plot[domain=-6:6] (2+\x/20,{-1.75+0.5/(1 + exp(-\x))-0.5});
    \node[neuron] (z4) at ( 3,-2) {$-{\bf B}^{(1)}$};
    \node[]     (plus) at ( 4,-2) {+};
    \node[]       (zo) at ( 5,-2) {};

    \draw[dashed] (-0.25,-3.5) rectangle (4.5,0.5);

    \draw[arrow] (0,-2) |- (B2);
    \draw[arrow] (z)    -- (z2);
    \draw[arrow] (y)    -- (y1);
    \draw[arrow] (f)    -- (f1);
    \draw[arrow] (z2)   -- (z3);
    \draw[arrow] (z3)   -- (z4);
    \draw[arrow] (z4)   -- (plus);
    \path[arrow] (z4)   -- (plus);
    \draw[arrow] (4, 0) -- (plus);
    \draw[arrow] (4,-3) -- (plus);
    \draw[arrow] (plus) -- (zo);
    \draw[arrow] (f1)   -- (fo);
    \draw[arrow] (plus) -- (zo);
    \draw[arrow] (y1)   -- (yo);
    \draw[arrow] (B2)   -- (plus);

\end{tikzpicture}
\end{center}
\caption{The ``encoding'' step for each patch can be conceptualised as recurrent neural network, consisting of matrix-vector multiplications and a softmax. The section within the dotted-lines represents each iteration of the variational updates of the latent variables. \label{fig:resnet}}
\end{figure}

In practice, this is iterated five times per patch for every full sweep through the image.
This specific number was chosen based on what was reported in \citet{khan2010variational}.
It may be worth noting that these variational updates of the latent variables consist only of matrix-vector multiplications, additions and a softmax.
The procedure can be conceptualised as a sort of ResNet, which we have attempted to illustrate in Fig. \ref{fig:resnet}.

\def\layersep{2.5cm}
\def\xpos{2}

\subsubsection{Decoding}

Once the expectation of the latent variables has been computed for the patches, the probabilistic label map can then be generated using $\hat{\tensor{W}}^{(2)}$ and $\hat{\bf M}^{(2)}$.
Slightly more accurate probabilities could be achieved by repeated sampling from ${\bf z}^* \sim \mathcal{N}(\hat{\bf z}^*,{\bf V}^*)$, but our approach simply reconstructs voxel probabilities using $\hat{\bf z}^*$.
\begin{align}
{\bf f}_i^{*(2)} \sim \Cat\left(\sigma\left(\hat{\bf W}_i^{(2)} \hat{\bf z}^* + \hat{\boldsymbol{\mu}}_i^{(2)}\right)\right)
\end{align}

\subsubsection{Registration with trained model}

Labellings achieved from the simple encoding-decoding model, while fast to compute, are of limited accuracy.
We note that our proposed model also allows a subject-specific template (see Fig. \ref{Fig:template_recon}) to be generated, such that
\begin{align}
{\bf f}_i^{*(1)} \sim \Cat\left(\sigma\left(\hat{\bf W}_i^{(1)} \hat{\bf z}^* + \hat{\boldsymbol{\mu}}_i^{(1)}\right)\right).
\end{align}

Because the trained model is able to generate synthetic template images with which new images can be aligned, this leads to a strategy that allows the alignment between any new image and the training data to be improved.
Higher labelling accuracies can be achieved by finessing the warps that align the images to label and the trained model.
For each subject's image data, this involves alternating between running the encoding and decoding model to generate a subject-specific template, and using this to refine the diffeomorphic alignment to achieve a closer match with the training data.
%%%%%%%%%%%%%%%%%%%%%%%%%%%%%%%%%%%%%%%%%%%%%%%%%%%%%%%%%%%%%%%%%%%%%%%%%%%%%

%%%%%%%%%%%%%%%%%%%%%%%%%%%%%%%%%%%%%%%%%%%%%%%%%%%%%%%%%%%%%%%%%%%%%%%%%%%%%
\section{Experiments \& Results}

Two experiments were performed.
The first involved assessing labelling accuracy compared with a ground truth based on manual annotations.
The second involved assessing the replicability of the labelling using different image contrasts.

\subsection{Datasets used}

We used the dataset from the \emph{MICCAI 2012 Grand Challenge and Workshop on Multi-Atlas Labeling}, which is available through \url{http://www.neuromorphometrics.com/2012_MICCAI_Challenge_Data.html}.
These data were provided for use in the MICCAI 2012 Grand Challenge and Workshop on Multi-Atlas Labeling \citep{landman2012miccai}.
The data is released under the Creative Commons Attribution-NonCommercial license (CC BY-NC) with no end date.
Original MRI scans are from OASIS (\url{https://www.oasis-brains.org/}).
Labelings were provided by Neuromorphometrics, Inc. (\url{http://Neuromorphometrics.com/}) under academic subscription.
The dataset consists of 35 manually labeled volumetric T1-weighted (T1w) MRI brain scans (1 mm isotropic resolution) from 30 unique subjects, with five of the subjects scanned twice.
The dataset is split into 15 training images and 20 testing images, where the testing images included those subjects who were scanned twice (the test-retest subjects).
We note that each of the images is an average of several rigidly-aligned face-stripped raw MRI scans.
Because missing voxels in the face-stripped scans were coded with a value of zero, it was necessary to erode the images to remove voxels containing an average of present and missing values.

We also used T1-, T2- and PD-weighted scans from the \emph{IXI} dataset (EPSRC GR/S21533/02), which is available from \url{https://brain-development.org/ixi-dataset/}.
This dataset is a collection of MR images from almost 600 normal, healthy subjects, which was collected on 1.5T and 3T MRI scanners at three different London hospitals.

\subsection{Tissue probability atlas}

Alignment and tissue segmentation of images prior to training and applying our proposed method used the approach described by \citet{brudfors2020flexible}, which extends ideas presented in \citet{blaiotta2018generative}.
These works combine diffeomorphic image registration and Gaussian mixture model based tissue classification within the same generative model, and also allow average shaped tissue probability maps to be computed.
The software is incorporated into SPM12 as the Multi-Brain (MB) toolbox.
Default settings were used throughout, except for the regularisation for the diffeomorphic registration, which was set to be higher than the default settings (``Shape Regularisation'' on the user interface was set to [0.0001 0.5 0.5 0.0 1.0]).

\begin{figure}
\begin{center}
\epsfig{file=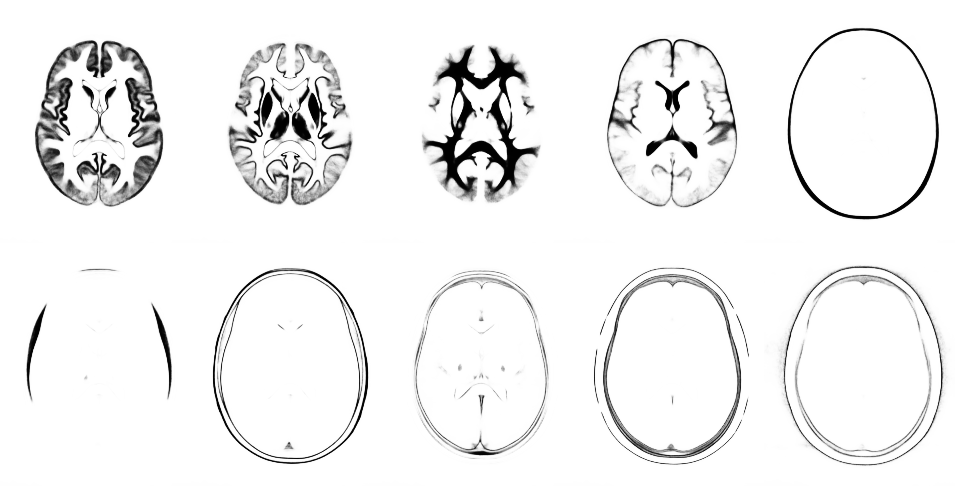,width=\textwidth}\\
\end{center}
\caption{
Tissue priors used for diffeomorphic alignment and tissue classification (not showing the background class).
The first three classes encode most of the tissues within the brain, and roughly encode pure grey matter, partial volume grey/white and pure white matter.
\label{Fig:tpm}}
\end{figure}

First of all, the \citet{brudfors2020flexible} approach was used to construct a tissue probability map from T2-weighted and PD-weighted scans of the first 64 subjects from the IXI dataset, along with the T1-weighted scans of the next 64 IXI subjects.
The 15 training subjects' scans from the MICCAI Challenge Dataset were also included in the template construction.
After merging several of the automatically identified tissue classes, the tissue probability map has  1 mm isotropic resolution, dimensions of 191$\times$243$\times$229 voxels and consists of 11 tissue types, three of which approximately corresponded with brain tissues.
This atlas of tissue priors (illustrated in Fig. \ref{Fig:tpm}) was used for all subsequent image alignment and tissue classification.

\subsection{Tuning settings on the MICCAI Challenge training dataset}

Scans from all 15 training subjects from the \emph{MICCAI 2012 Grand Challenge and Workshop on Multi-Atlas Labeling} dataset were aligned with the tissue prior atlas (described previously) and the brains segmented into three tissue classes using the method of \citet{brudfors2020flexible}.
To facilitate training, warped versions of the tissue classes and labels were generated for the training subjects at 1 mm isotropic resolution.
Warped tissue class images were saved in NIfTI format, whereas a custom sparse matrix format was used to encode the warped labels.

Subsequently, various settings for training the proposed method were then tuned using data from these 15 subjects, whereby the first 10 were used for model training, with the remaining five used for validation.

Following training, labelling of the five validation scans was done with either:
\begin{itemize}
\item No additional diffeomorphic registration.
\item Four Gauss-Newton updates of additional diffeomorphic registration, which involves alternating between estimating the latent variables and updating the alignment.
The registration regularisation was one quarter of that used for the initial registration/segmentation of the data.
\end{itemize}
Because label predictions were made in normalised space, the label probabilities were warped back to match the original image volumes using partial volume interpolation \citep{maes1997multimodality}, where they were converted into categorical data by assigning the most probable label at each voxel.
Accuracy was assessed by the average Dice-S\/orensen Coefficient (DSC), measuring the overlap between the ground truth and our predictions.

It was not feasible to explore all possible model settings via a full grid search, so we selected a few choice settings and examined the DSC that these achieved on the validation set.
Because training is slower when augmentation is used, the initial tuning was done without augmentation.
Four outer iterations were used during each training attempt.

Because relatively small regions had proven successful in previous label propagation works, we chose to use patch sizes of 4$\times$4$\times$4 voxels.
The model accounts for the covariance among latent variables in neighbouring patches, so it is able to induce locally correlated behaviour across neighbouring patches.
If the model can benefit from weights in neighbouring patches always varying in unison, then this will be exploited.
Therefore, we did not explore patch sizes and instead focussed on those settings that control the behaviour of the conditional random field.

Initially, the main settings that were varied were those controlling the Wishart prior on the conditional random field ($\nu_0$ and $v_0$, where $\boldsymbol\Psi_0^{-1} = \mathbf{I}\nu_0 v_0$).
Three values for $\nu_0$ were used: the first was a value of 1.0, which encodes an improper prior; the second used the least informative proper prior in a way that varied according to how the model was pruned (named ``var'' in Table \ref{table:validation_DSC}); the third was a relatively uninformative proper prior based on $\nu_0 = 7K-0.9$.
Three different values for $v_0$ were also explored.
For these experiments, the maximum number of basis functions $K$ was set to nine, although one run involved training with $K$=0 to serve as a majority-voting baseline.

Once suitable Wishart prior settings were identified, the next step was to continue the tuning using data that have been augmented by translating by up to 1.5 voxels. This scaled the amount of training data by a factor of 19, leading to a concomitant increase in training times.
We considered that $K$=16 would be a reasonable maximum number of latent variables to use for each patch.
These experiments varied the standard deviation of the Gaussian weighting used for augmentation.
A final run involved augmenting by translating by up to 3 voxels during training, which increased the training time by about a factor of 123.

\begin{table}[h!]
\centering
\caption{Training/validation DSC with different settings.}
\label{table:validation_DSC}
\begin{adjustbox}{max width=\textwidth}
    \begin{tabular}{|c|c|c|c|c||c|c|c||c|c|c|}
    \hline
    \multicolumn{5}{|c||}{Settings} &
    \multicolumn{3}{c||}{DSC - no registration} &
    \multicolumn{3}{c|}{DSC - with registration} \\
    \hline
    $\nu_0$ & {$v_0$} & $K$ & r & sd & Overall & Non-cort. & Cortical  & Overall & Non-cort. & Cortical\\
    \hline
    \hline
    N/A  & N/A  &  0 &  0.0  & 0.0 &  0.7008 & 0.8215 & 0.6564  & 0.7260 & 0.8341 & 0.6862 \\
    \hline
    1.0  & 0.3  &  9 &  0.0  & 0.0 &  0.7264 & 0.8370 & 0.6858  & 0.7423 & 0.8429 & 0.7053 \\
    1.0  & 1.0  &  9 &  0.0  & 0.0 &  0.7277 & 0.8374 & 0.6874  & 0.7427 & 0.8427 & 0.7059 \\
    1.0  & 3.0  &  9 &  0.0  & 0.0 &  0.7284 & 0.8373 & 0.6883  & 0.7426 & 0.8428 & 0.7058 \\
    var  & 0.3  &  9 &  0.0  & 0.0 &  0.7294 & 0.8385 & 0.6894  & 0.7440 & 0.8442 & 0.7072 \\
    var  & 1.0  &  9 &  0.0  & 0.0 &  0.7302 & 0.8385 & 0.6904  & 0.7435 & 0.8440 & 0.7066 \\
    var  & 3.0  &  9 &  0.0  & 0.0 &  0.7301 & 0.8386 & 0.6902  & 0.7424 & 0.8436 & 0.7052 \\
   62.1  & 0.3  &  9 &  0.0  & 0.0 &  0.7285 & 0.8366 & 0.6889  & 0.7435 & 0.8426 & 0.7071 \\
   62.1  & 1.0  &  9 &  0.0  & 0.0 &  0.7295 & 0.8369 & 0.6900  & 0.7432 & 0.8428 & 0.7065 \\
   62.1  & 3.0  &  9 &  0.0  & 0.0 &  0.7294 & 0.8374 & 0.6898  & 0.7421 & 0.8430 & 0.7051 \\
    \hline
    var  & 1.0  & 16 &  1.5  & 1.0 &  0.7405 & 0.8419 & 0.7033  & 0.7504 & 0.8453 & 0.7156 \\
    var  & 1.0  & 16 &  1.5  & 3.0 &  0.7408 & 0.8416 & 0.7037  & 0.7506 & 0.8451 & 0.7158 \\
    var  & 1.0  & 16 &  1.5  &10.0 &  0.7408 & 0.8416 & 0.7037  & 0.7506 & 0.8450 & 0.7159 \\
    var  & 1.0  & 16 &  3.0  & 3.0 &  0.7438 & 0.8392 & 0.7087  & 0.7526 & 0.8429 & 0.7194 \\
    \hline
    \end{tabular}
\end{adjustbox}
\end{table}

DSC scores are presented in Table \ref{table:validation_DSC}.
Varying the settings of the Wishart prior made relatively little difference to the DSC, although the variable setting for $\nu_0$ with $v_0$=1.0 was the most effective without additional registration.
Augmentation and using additional registration gave greater improvements.
Although using the 3 voxel radius augmentation led to less accurate labelling of non-cortical structures, it gave the best overall DSC.

\begin{figure}
\begin{center}
\epsfig{file=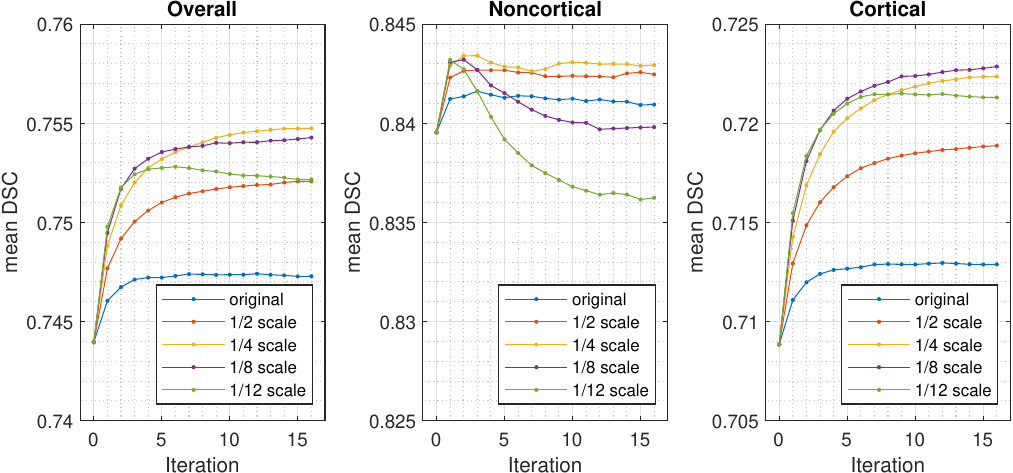,width=\textwidth}\\
\end{center}
\caption{
DSC for validation set using different numbers of registration iterations and degrees of regularisation, relative to that used for the initial registration and segmentation.
\label{Fig:registration_plot}}
\end{figure}

The effects of the number of Gauss-Newton iterations used for the registration refinement was assessed, as well as the amount of regularisation.
This used the model trained with 3 voxel radius augmentation, and involved scaling the regularisation used for the initial registration by various different amounts.
For each registration iteration, re-estimation of latent variables used 10 outer iterations (sweeps over the patches) and 16 inner iterations (recomputing latent variables at each patch), which was more than was used for the results in Table \ref{table:validation_DSC} (9 and 5 respectively).
The resulting DSC are plotted in Fig. \ref{Fig:registration_plot}.

\subsection{Accuracy on MICCAI Challenge test dataset}

The FIL model was then re-trained on all 15 training subjects, using a minimally informative, but proper Wishart prior, with $v_0$=1.0.
An augmentation search radius of 3 voxels was used with a Gaussian weighting standard deviation of 2.0 voxels.
Patch sizes were 4$\times$4$\times$4 voxels, and four outer iterations were used for model training.
Up to $K$=24 basis functions were available to encode each patch, although no more than 15 were needed for any patch after the automatic pruning.

The test subjects were automatically labelled, using six Gauss-Newton iterations for the additional registration.
As for the MICCAI challenge, the DSC was computed over all 20 target scans, with results presented as average DSC for cortical labels, non-cortical labels and the overall average.
Table \ref{table:1} shows the accuracies from the proposed method, alongside previous results from the literature where the same MICCAI challenge data were used.
We also trained the model without any basis functions (i.e., $K$=0) or augmentation, which gave majority voting predictions for the spatially normalised data.
These results are also presented in Table \ref{table:1}
and show that the proposed FIL method increases the overall overlap from a majority voting baseline by about 2.7\% (1.5\% and 3.2\% for non-cortical and cortical regions respectively).
For comparison, the PICSL joint label fusion method \citep{wang2013multi} gave similar improvements over majority voting (3.1\% overall, 2.9\% non-cortical and 3.1\% cortical), although the authors achieved additional DSC increases by including their corrective learning step (additional 1.4\%, 1.1\% and 1.5\% respectively).
We note that the regularisation used for the registration was quite high, and a higher DSC baseline may have been achieved if this regularisation was lower.

There were 25 entries to the original MICCAI challenge, and the top five entries are also shown in Table \ref{table:1}. 
These were PICSL-BC \citep{wang2012combined}, NonLocalSTAPLE \citep{asman2012multi}, MALP-EM \citep{ledig2012segmentation}, PICSL-Joint \citep{wang2012combined} (same as PICSL-BC, but without the corrective learning \citep{wang2011learning}) and MAPER \citep{heckemann2012multi}.
Since then, a number of other papers have reported accuracies on these data that were obtained using other methods, so the table also includes several of those results.

While the average DSC from our proposed FIL method were not as high as those from the top performing methods, they would still have achieved fifth position on the leaderboard of the MICCAI challenge. 
Most of the challenge entries required each test scan to have been registered pairwise with all of the 15 training scans, but our proposed method used a single nonlinear registration for each test scan to the tissue probability template, followed by iterative refinement of the registration, which saves a considerable amount of time.
After a single tissue classification and registration (taking about 23 minutes per subject), the labelling itself took about 24 minutes (3.25 minutes without the additional registration) to label each volumetric T1w scan. The laptop computer used in this work is an ASUS ZenBook 14 UX434 with 8 GB of RAM and an AMD Ryzen 5 3500U processor.
A GPU implementation would likely lead to much better performance.

\begin{figure}
    \includegraphics[width=\textwidth]{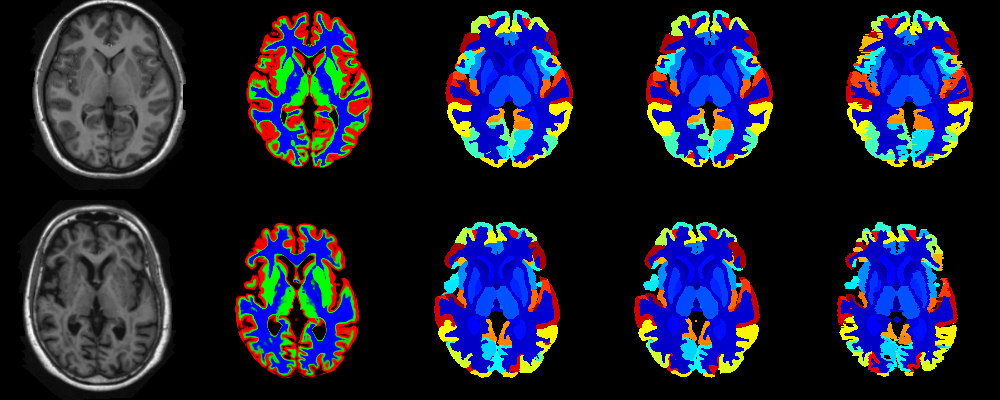}
    \caption{Best case (scan 1038, top row) and worst case (scan 1128, bottom row) labellings from the fusion challenge dataset. The figure shows the original T1w scan (first column), the tissue classes (second column) identified using the method of \citet{brudfors2020flexible}, the majority voting labelling (third column), the predicted labels using the proposed FIL method (fourth column) and the manually defined ground truth labels (fifth column).\label{fig:labelmap}}
\end{figure}

\begin{figure}
    \includegraphics[width=\textwidth]{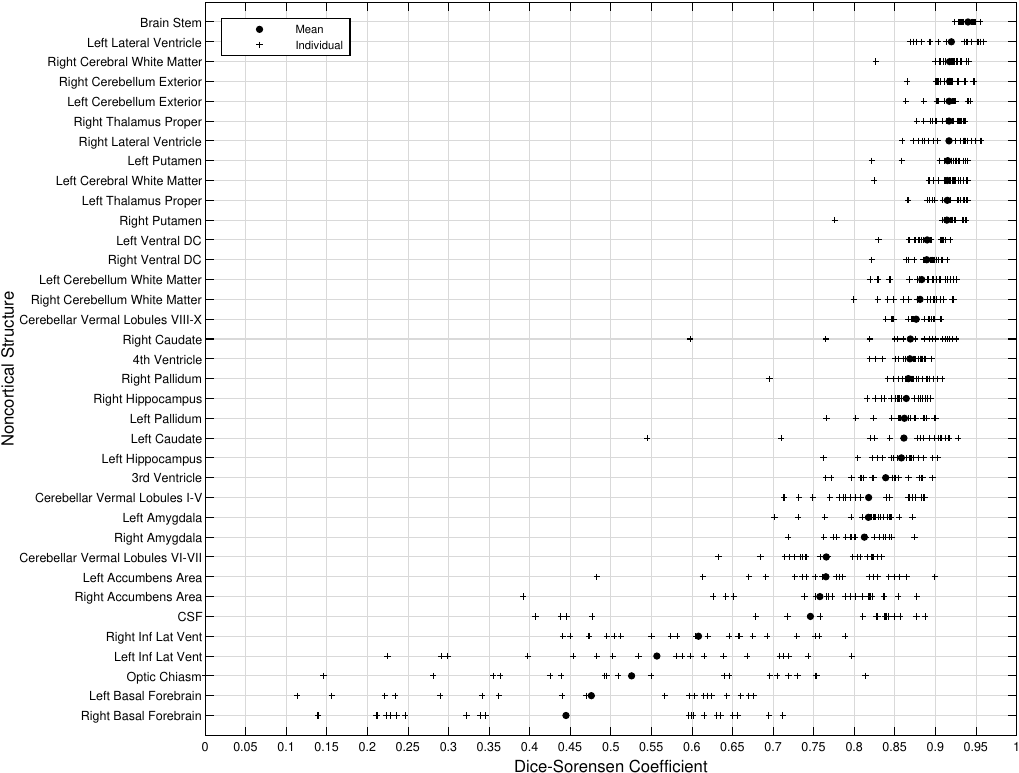}
    \caption{DSC for non-cortical brain regions (fusion challenge data). Plots show individual DSC ($+$) as well as average DSC ($\bullet$) across subjects for each brain region.\label{fig:MICCAI2012-noncortical}}
\end{figure}

\begin{figure}
    \includegraphics[width=1.1\textwidth]{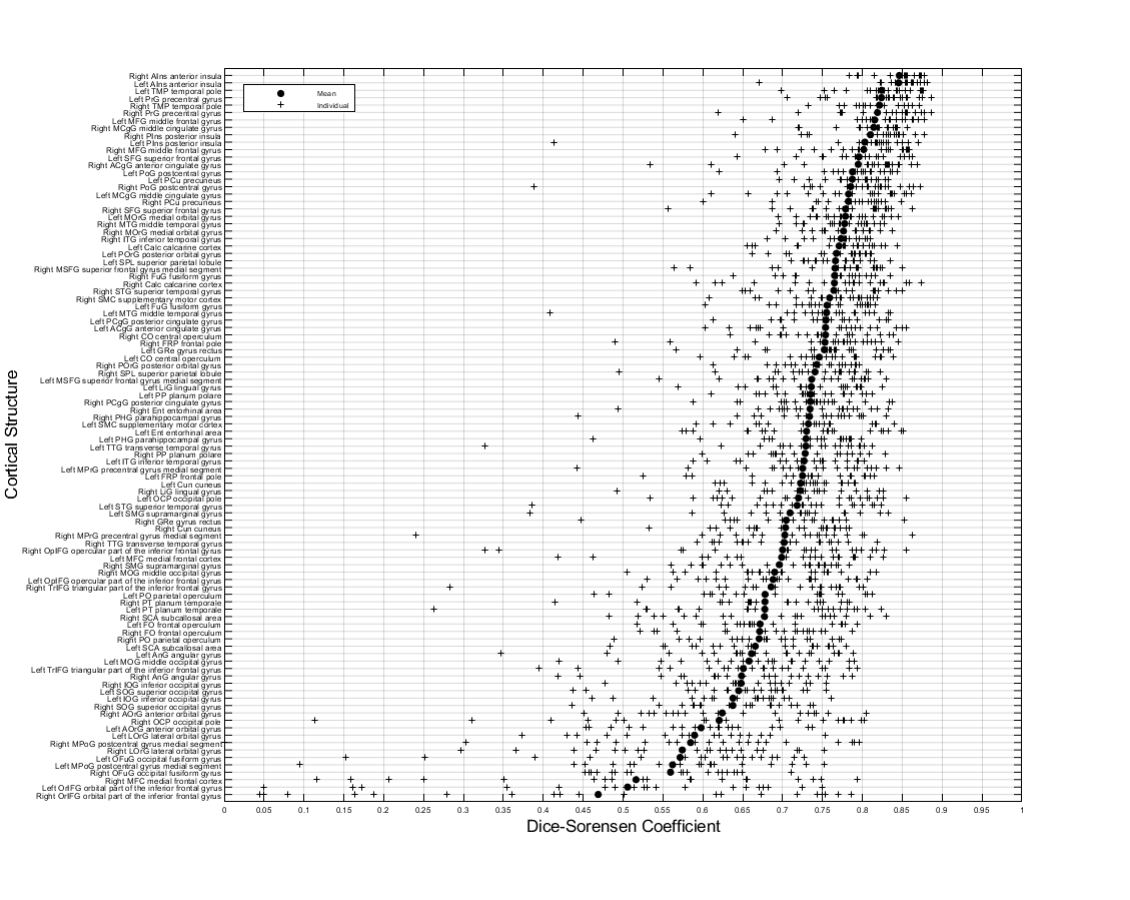}
    \caption{DSC for cortical brain regions (fusion challenge data). Plots show individual DSC ($+$) as well as average DSC ($\bullet$) across subjects for each label.\label{fig:MICCAI2012-cortical}}
\end{figure}

Fig. \ref{fig:labelmap} illustrates the predicted and ground truth labellings, along with their tissue classifications, for the scans with the highest and lowest average DSC.
As can be seen, the white matter hyperintensities in the scan with the lowest average DSC led to less accurate tissue classification, which in turn resulted in less accurate FIL labellings.  
The DSC (all individuals and mean) for non-cortical and cortical brain regions are shown in Figs. \ref{fig:MICCAI2012-noncortical} and \ref{fig:MICCAI2012-cortical} respectively.

\begin{table}[h!]
\centering
\caption{Average Dice-S\/orenson coefficients (DSC) across testing subjects using the proposed method in comparison with published methods, including the top three approaches from the MICCAI workshop.}
\label{table:1}
\begin{adjustbox}{max width=\textwidth}
    \begin{tabular}{*{14}{|l}|}
    \hline
    Method & overall   & non-cortical  & cortical  \\
    \hline
    \hline
    MICCAI-2012:  1. PICSL-BC \citep{wang2012combined}     & 0.765 & 0.838 & 0.739 \\
    MICCAI-2012:  2. NonLocalSTAPLE \citep{asman2012multi} & 0.758 & 0.830 & 0.732 \\
    MICCAI-2012:  3. MALP-EM \citep{ledig2012segmentation} & 0.758 & 0.825 & 0.733 \\
    MICCAI-2012:  4. PICSL-Joint \citep{wang2012combined}  & 0.750 & 0.827 & 0.722 \\
    MICCAI-2012:  5. MAPER \citep{heckemann2012multi}      & 0.741 & 0.814 & 0.714 \\
    MICCAI-2012: 25. Last-place entry                      & 0.711 & 0.786 & 0.683 \\
    \hline
    Joint label fusion + corrective learning \citep{wang2013multi} & 0.771 & 0.836 & 0.747 \\
    Random Forest \citep{zikic2014encoding}            & 0.728 & 0.805 & 0.699 \\
    CNN \citep{de2015deep}                             & 0.725 & -     & -     \\
    CNN \citep{moeskops2016automatic}                  & 0.735 & 0.785 & 0.717 \\
    CNN \citep{mehta2017brainsegnet}                   & 0.743 & 0.805 & 0.720 \\
    Transfer learning FCN \citep{huo20193d}            & 0.776 & -     & -     \\
    \hline
    Majority voting (proposed)                         & 0.717 & 0.804 & 0.684 \\
    FIL (proposed)                                     & 0.744 & 0.819 & 0.716 \\
    \hline
\end{tabular}
\end{adjustbox}
\end{table}

To better understand the upper limit of the accuracies that may be achieved, the test-retest subjects were coregistered together using SPM12's implementation of normalised mutual information coregistration \citep{studholme1999overlap} and the label maps that had been manually defined on the second scans were resliced to match those of the first scans using partial volume interpolation.
DSC was computed between the first scan labels and the resliced second scan labels and the overall average was found to be 0.816 (0.846 for non-cortical and 0.805 for cortical).
Similarly, the hard labels generated by the FIL method for the second scans were resliced, and the overall average DSC was 0.933 (0.932 for non-cortical and 0.934 for cortical).
This shows higher test-retest reliability for the FIL method compared with with manual labelling. 

\subsection{Replicability Under Domain Shift}

This section assesses the replicability of the proposed label propagation method, by computing DSC between labellings computed from T1w scans, versus those obtained from jointly using T2w and PDw scans of the same subjects.
The last 10 subjects for each of the three scanning sites within the \emph{IXI} dataset were used for this work (Guys Hospital: IXI639 -- IXI662; Hammersmith Hospital: IXI632 -- IXI646; Institute of Psychiatry: IXI553 -- IXI596).
The T2w and PDw scans were rigidly aligned with the T1w scans of each subject using normalised mutual information \citep{studholme1999overlap}.

\begin{figure}
    \includegraphics[width=\textwidth]{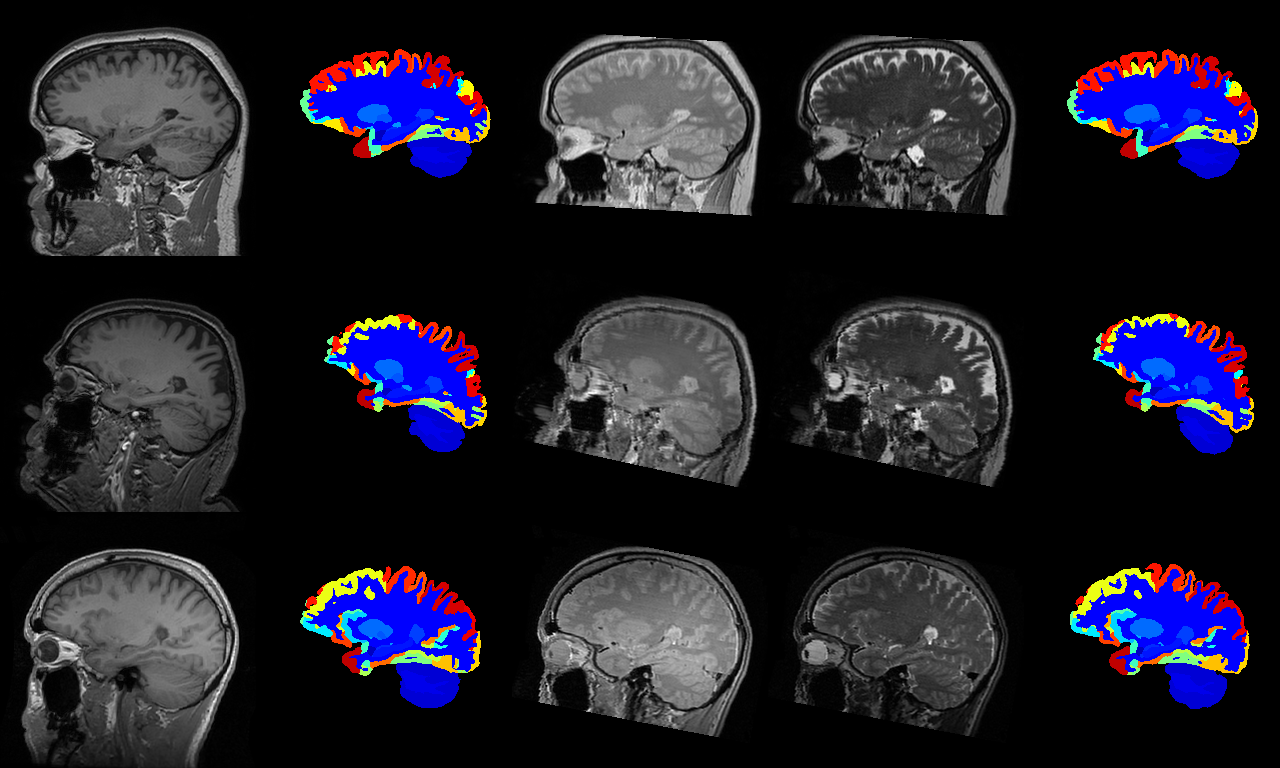}
    \caption{Example labellings of different modalities from three different scanners.  First column: T1-weighted scan. Second column: Labelled T1-weighted scan. Third and fourth columns: PD- and T2-weighted scans. Fifth column: Labelled PD- and T2-weighted scans.  First row: Guys Hospital; Second row: Hammersmith Hospital; Third row: Institute of Psychiatry. \label{fig:ixipic}}
\end{figure}

The scans were subsequently processed as described previously, and labelled using the FIL model trained on the \emph{MICCAI 2012 Workshop on Multi-Atlas Labeling} dataset, as described in the previous section.
No ground truth is available for these data so we simply assess the DSC between the two sets of labellings, so that the DSC measures how similar a labelling obtained from a subject's T1w scan is to the labelling obtained from their T2w and PDw scans.
Example labellings are shown in Fig. \ref{fig:ixipic} and average DSC from scans from the three different sites are presented in Table \ref{table:2}.

\begin{table}[h!]
\centering
\caption{Average DSC between FIL labelling of T1w scans versus jointly labelling T2w and PDw scans.
    The average DSC was computed for data from each site, as well as the overall average DSC.}
\label{table:2}
\begin{adjustbox}{max width=\textwidth}
    \begin{tabular}{*{14}{|l}}
    \hline
    Method & overall   & non-cortical  & cortical  \\
    \hline
    \hline
    FIL: Guys Hospital (Philips 1.5T)      & 0.835 & 0.848 & 0.830 \\
    FIL: Hammersmith Hospital (Philips 3T) & 0.776 & 0.754 & 0.784 \\
    FIL: Institute of Psychiatry (GE 1.5T) & 0.745 & 0.743 & 0.745 \\
    FIL: Overall                           & 0.785 & 0.782 & 0.786 \\
    \hline
\end{tabular}
\end{adjustbox}
\end{table}

As can be seen from Table \ref{table:2}, the overall DSC of 0.785 indicates that the two sets of labellings are similar, although there was considerable systematic variability in this similarity across the different sites.
Visual inspection of the \emph{IXI} scans (not shown) suggests that better grey-white contrast in the T2w and PDw scans in the Guys Hospital data led to more consistent tissue segmentations, which in turn led to greater labelling consistency.
These results suggest that the FIL method is able to generalise well to scans that have sufficient contrast to achieve good tissue classification.
%%%%%%%%%%%%%%%%%%%%%%%%%%%%%%%%%%%%%%%%%%%%%%%%%%%%%%%%%%%%%%%%%%%%%%%%%%%%%

%%%%%%%%%%%%%%%%%%%%%%%%%%%%%%%%%%%%%%%%%%%%%%%%%%%%%%%%%%%%%%%%%%%%%%%%%%%%%
\section{Discussion}

We have proposed a patch-based, variational Bayesian model for label map prediction using aligned tissue segmented MR scans from SPM12.
We computed the DSC for overall brain, cortical and non-cortical regions, which we compared with the MICCAI 2012 challenge leader table.
We also assessed the replicability of labelling MRI data acquired with different image contrasts.

Applying the basic method to a new scan only uses a single image registration, which is subsequently refined, to align with an average shaped template, rather than several separate registration steps to align with all the training scans.
We also note that the proposed method is applied to automatically identified tissue classes, which we hope will enable it to generalise better to a broader range of new image types.

In its current form, our proposed labelling approach is not as accurate as some other approaches when applied to scans with the same MRI contrast as the training images.
One likely reason for this is that the tissue classification itself may be the limiting factor, as it uses an atlas of tissue priors based only on spatial warping.
Because our proposed model is generative, it may be better able to encode priors that could be used for tissue classification, overcoming the limitations of purely warping-based priors.
This would also increase compatibility between the tissue classification method and the labelling method, as both would use the same underlying model.
However, this may come with the cost of making it more difficult to formulate the training so that the results generalise to scans with a wide variety of MR contrasts.
We leave these explorations to future work.

In addition to experimenting with different model settings, there are a number of other areas that could lead to potential improvement.
As suggested in \citet{sabuncu2010generative}, the simple data augmentation approach could probably be improved by augmenting based on the uncertainty of the image registration \citep{simpson2012ensemble,iglesias2013improved,wang2018efficient}, which would effectively ``integrate out'' this source of uncertainty.
Such an approach would also need to consider the expected uncertainty with which target images could be aligned.
While it benefited cortical segmentation, we found that our simple augmentation strategy often decreased DSC for many non-cortical structures, where registration uncertainty was lower.
This supports the idea that it would be more effective to augment according to this uncertainty.

The proposed method is generative, so we speculate that this may make it easier to extend to do semi-supervised and multi-task learning.
For example, it might be possible to achieve greater accuracy by training using additional data that does not have manually defined labels associated with it.
Although we know that this would allow the model parameters used for encoding (i.e., $\hat{\tensor{W}}^{(1)}$, ${\bf M}^{(1)}$, $\nu$ and $\boldsymbol\Psi$) to be more accurately characterised, we can currently only speculate on whether this would translate into more accurate labelling.
Similarly, the model could be learned by combining sets of annotations defined using different labelling protocols.
While this could be another approach for estimating more accurate encoding parameters, we do not yet know whether encoding many different aspects of images using the same sets of latent variables would improve overall performance. 
%%%%%%%%%%%%%%%%%%%%%%%%%%%%%%%%%%%%%%%%%%%%%%%%%%%%%%%%%%%%%%%%%%%%%%%%%%%%%

%%%%%%%%%%%%%%%%%%%%%%%%%%%%%%%%%%%%%%%%%%%%%%%%%%%%%%%%%%%%%%%%%%%%%%%%%%%%%
\section*{Conflict of Interest Statement}

The authors declare that the research was conducted in the absence of any commercial or financial relationships that could be construed as a potential conflict of interest.

\section*{Author Contributions}

YY contributed to the code, performed the evaluations and helped draft the manuscript.
YB and MB provided important advice and helped draft the manuscript.
JA formulated the equations, contributed to the code, ran some evaluations and helped draft the manuscript.

\section*{Funding}

This project has received funding from the European Union’s Horizon 2020 Research and Innovation Programme under Grant Agreement No. 785907 (Human Brain Project SGA2).
This research was funded in whole, or in part, by the Wellcome Trust [203147/Z/16/Z]. For the purpose of Open Access, the author has applied a CC BY public copyright licence to any Author Accepted Manuscript version arising from this submission.

\section*{Data Availability Statement}

Code for the Factorisation-based Image Labelling and Multi-Brain toolbox is available from \url{https://github.com/WCHN/Label-Training/} and \url{https://github.com/WTCN-computational-anatomy-group/mb} respectively, and will eventually be developed as a toolbox for the SPM package (available from \url{https://www.fil.ion.ucl.ac.uk/spm/software/spm12/}).
The MICCAI 2012 workshop on Multi-Atlas Labeling dataset is available from \url{http://www.neuromorphometrics.com/2012_MICCAI_Challenge_Data.html}.
These data were provided for use in the MICCAI 2012 Grand Challenge and Workshop on Multi-Atlas Labeling \citep{landman2012miccai}.
The data is released under the Creative Commons Attribution-NonCommercial license (CC BY-NC) with no end date.
Original MRI scans are from OASIS (\url{https://www.oasis-brains.org/}).  Labelings were provided by Neuromorphometrics, Inc. (\url{http://Neuromorphometrics.com/}) under academic subscription.
The \emph{IXI} dataset (EPSRC GR/S21533/02) is available from \url{https://brain-development.org/ixi-dataset/}.

%%%%%%%%%%%%%%%%%%%%%%%%%%%%%%%%%%%%%%%%%%%%%%%%%%%%%%%%%%%%%%%%%%%%%%%%%%%%%

\bibliography{References}

%%%%%%%%%%%%%%%%%%%%%%%%%%%%%%%%%%%%%%%%%%%%%%%%%%%%%%%%%%%%%%%%%%%%%%%%%%%%%
\end{document}